\documentclass{article}

\usepackage{graphicx} 

\usepackage[utf8]{inputenc} 
\usepackage[T1]{fontenc}    
\usepackage{hyperref}       
\usepackage{url}            
\usepackage{booktabs}       
\usepackage{amsfonts}       
\usepackage{nicefrac}       
\usepackage{microtype}      
\usepackage{xcolor}         
\usepackage{amsmath}         
\usepackage{graphicx}
\usepackage{makecell}
\usepackage{amssymb} 
\usepackage[utf8]{inputenc}
\usepackage{caption}  

\title{TSDiT: Traffic Scene Diffusion Models With Transformers}

\author{
  Chen Yang \\
  Cardiff University \\
  \texttt{rickyyang1997@163.com} \\
  \and
  Tianyu Shi \\
  University of Toronto \\
  \texttt{ty.shi@mail.utoronto.ca} \\
}

\begin{document}

\maketitle
\begin{abstract}
  In this paper, we introduce a novel approach to trajectory generation for autonomous driving, combining the strengths of Diffusion models and Transformers. First, we use the historical trajectory data for efficient preprocessing and generate action latent using a diffusion model with DiT(Diffusion with Transformers) Blocks to increase scene diversity and stochasticity of agent actions. Then, we combine action latent, historical trajectories and HD Map features and put them into different transformer blocks. Finally, we use a trajectory decoder to generate future trajectories of agents in the traffic scene. The method exhibits superior performance in generating smooth turning trajectories, enhancing the model's capability to fit complex steering patterns. The experimental results demonstrate the effectiveness of our method in producing realistic and diverse trajectories, showcasing its potential for application in autonomous vehicle navigation systems. 
\end{abstract}

\begin{figure}[htbp]
    \centering
    \includegraphics[width=0.8\textwidth]{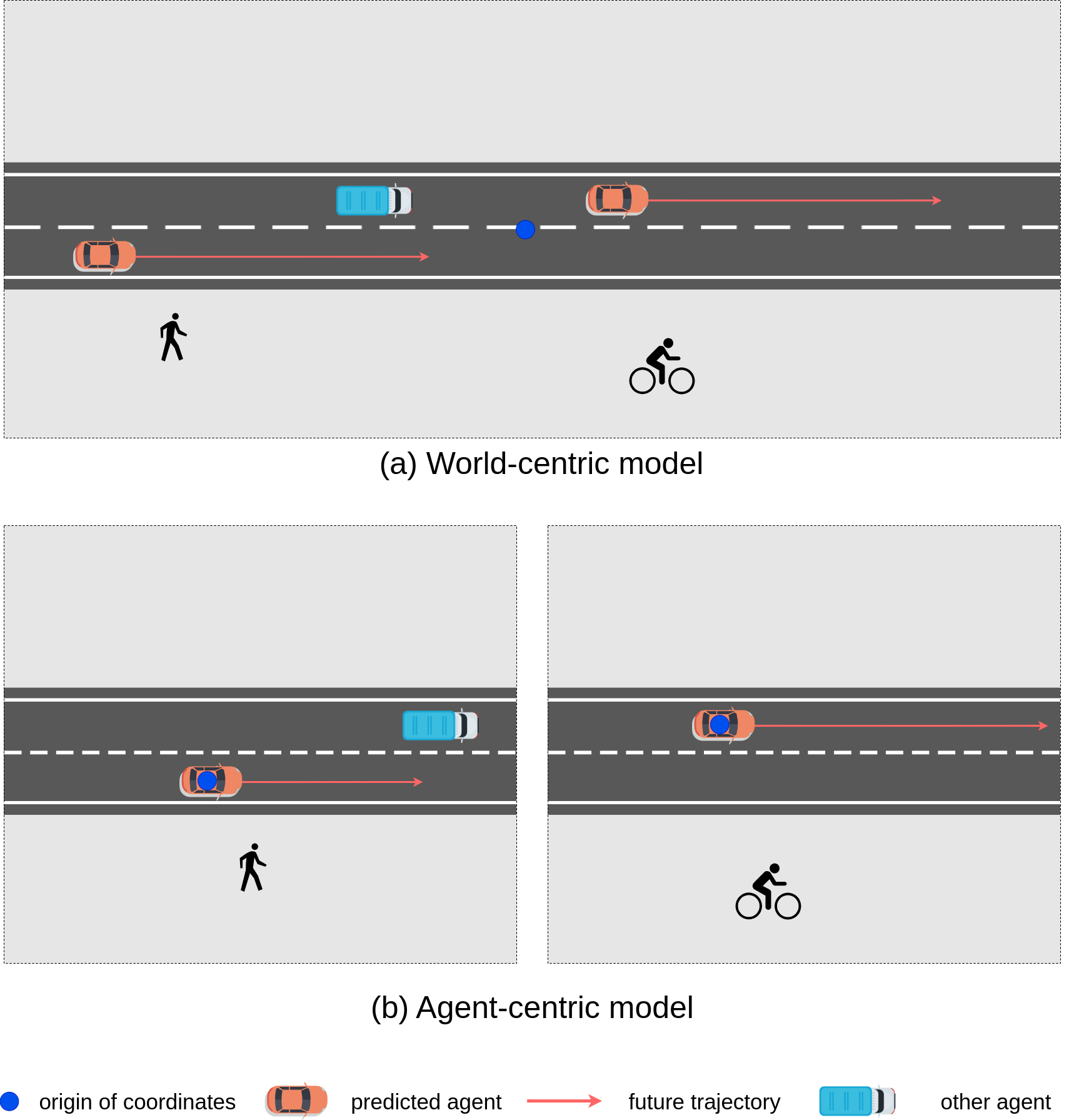}
    \caption{
        World-centred model and Agent-centred model: (a) World-centred model: firstly, a "world centre" is defined, and then the map information and the positions of the agents in the traffic scene are transformed from absolute coordinates to frenet coordinates with the "world centre" as the origin. In the same traffic scenario, all features have the same coordinate origin, so the world-centred model can simultaneously output the future trajectories of all agents in the traffic scenario in one inference. (b) Agent-centred model: the map information and other agents' positions in the traffic scene are transformed from absolute coordinates to frenet coordinates with each agent as the origin, which means if there are $N$ agents in a traffic scene, the features of $N$ agents need to be input and infer $N$ times to obtain their trajectories.
    }
    \label{fig:difference_gene_traj_and_pred}
\end{figure}

\section{Introduction}
Simulators play a crucial role in advancing autonomous driving systems (ADS), providing a cost-effective and efficient means to assess the performance of driving algorithms. Unlike vehicle testing, simulators enable comparative experiments using consistent traffic scenarios. However, a significant challenge faced by most simulations lies in accurately replicating real-world traffic behaviors. Current simulation methods ~\cite{dosovitskiy2017carla,behrisch2011sumo} primarily involve two approaches: first, replaying recorded real traffic scenes, which lacks the ability to alter traffic participant behavior based on different autonomous driving algorithms; and second, defining obstacle behavior using rules, which fails to precisely emulate real-world obstacle movement. These limitations underscore the need for more sophisticated simulation techniques to bridge the gap between simulated scenarios and real-world driving conditions.

Recently, generative models have gained prominence for their ability to learn and replicate intricate data distributions, finding applications in diverse domains like image, text, and video generation~\cite{yang2022diffusion}. To address the challenge of generating realistic simulation environments, numerous learning-based models have been proposed. These models leverage "real world" data distributions and utilize the current "traffic state" to generate obstacle actions.  This study explores the potential of diffusion models in generating high-quality traffic scenes compared to other generative models.

In this study, we introduce a novel traffic scene generation model named Traffic Scene Diffusion Models With Transformers(TSDiT). Our model consists of two integral components: a diffusion model that produces "action distributions" derived from traffic features, and a trajectory decoder responsible for generating trajectories based on these "action latent". More importantly, the existing methods in the Sim Agent Challenge mainly use the idea of agent prediction to achieve traffic scenario generation~\cite{montali2023waymo}, which has two significant disadvantages: 1. input features need to be transformed into frenet coordinates centred on each agent during the data preprocessing stage 2. only one agent's trajectory can be output per model inference. To address these issues, we have introduced a new data preprocessing approach and model structure in TSDiT, which we refer to as "world-centric model", as shown in Figure~\ref{fig:difference_gene_traj_and_pred}. Our primary contributions to this research can be highlighted as follows:

\begin{itemize}
\item We developed a "world-centric model" where the model inputs are not centred on each agent, but rather the actions of each agent in the traffic scenario are generated by inputting "global information".
\item We compare our model with other baselines on the Waymo motion prediction dataset and show the advantages of our model for traffic scenario generation.
\end{itemize}

\section{Related Work}
\subsection{Sim Agents Challenge}
The enhancement of autonomous driving system development efficiency relies substantially on a realistic interactive autonomous driving simulator capable of swiftly evaluating autonomous driving algorithms. In response to this need, the Waymo Open Sim Agents Challenge (WOSAC) was introduced~\cite{montali2023waymo}, aiming to foster the development of realistic simulators. WOSAC incorporates various dimensional metrics, such as kinematic metrics, object interaction metrics, and map-based metrics, to rigorously assess simulators in quantitative and qualitative terms. To further align the simulator's outcomes with real-world scenarios, WOSAC has introduced metrics like Average Displacement Error (ADE) and Final Displacement Error (FDE). These metrics serve to bridge the simulation-reality gap and enhance the fidelity of simulation results.


\subsection{Diffusion Model for Driving Scene Generation }

Proonovost et al \cite{pronovost2023generating} proposed "Scene Diffusion," a novel traffic scene generation model. It combines an autoencoder and a diffusion model, leveraging latent diffusion and object detection to directly produce bounding boxes for agents. The key contributions include an innovative end-to-end architecture and an evaluation of the model's ability to generalize across different geographical regions, enhancing the realism and adaptability of the generated driving scenes.  Zhong et al ~\cite{zhong2023guided} introduces CTG, a conditional diffusion model enabling controllable and realistic traffic simulation for autonomous vehicle development. It fills a gap by allowing users to control trajectory properties at test time while enforcing realism through guided diffusion modeling, leveraging signal temporal logic. Evaluation of the nuScenes dataset demonstrates CTG's effectiveness in balancing controllability and realism, showcasing improvements over strong baselines. Furthermore, Zhong et al \cite{zhong2023language} introduce CTG++, a scene-level conditional diffusion model for realistic and controllable traffic simulation in autonomous vehicle development. The model incorporates a spatio-temporal transformer backbone to generate lifelike traffic scenarios. Utilizing a large language model, it translates user queries into loss functions, guiding the diffusion model to produce simulations that align with specific instructions. Comprehensive evaluations demonstrate the effectiveness of CTG++ in generating realistic and query-compliant traffic simulations.

\section{"World-centric" Scene Understanding}
Traffic scenarios consist of multimodal data such as road information, traffic light status, predicted agent history, other agent history \cite{nayakanti2023wayformer}. In this section, we detail how these features are represented in our setup. For readability, we define the following notation: A denotes all the agents of the traffic scenario, $A_{p}$ indicates the number of agents to be predicted, $A_{o}$ indicates other agents in the traffic scenario except $A_{p}$, $T_{h}$ denotes the number of historical time steps considered, $T_{f}$ denotes the number of future time steps to be predicted, $L$ denotes all lane lines in a traffic scenario, $P$ denotes a point on a lane line $L$, $T_{tl}$ denotes traffic lights in a traffic scenario, $D$ denotes the feature dimension.

\subsection{Agent Classification}
In order to achieve the "world-centred model", we classify the agents in the traffic scenario into predicted agents and other agents, which are the agents provided in Sim Agents that need to be predicted, while other agents are defined as the agents in the traffic scenario that may interact with predicted agents in addition to predicted agents. Other agents are defined as the agents in the traffic scenario that may interact with predicted agents in addition to predicted agents.

\subsection{Agent History and Feature}
For predicted agents and other agents, the input features are the same which contain a series of past agent movement states $[A, T_{h}-1, D_{m}]$ and agent features $[A, 1, D_{a}]$. For the movement state , We use the difference between the current time step and the previous time step for features such as: position, angle, velocity. For the agent features, we fuse the agent location, agent size and agent type for the current time step.

\subsection{Traffic Light Feature}
For each traffic scenario s, traffic light information $[T_{tl}, T_{h}, D_{t}]$ contains the positions and signals of all traffic signals in a traffic scenario at historical time steps. For each traffic signal point $t_{tl} \in T_{tl}$, it represents the one hot encoding of traffic light signals and the position of traffic lights at each time step.

\subsection{Map Feature}
The map feature $[1, L, P, D_{m}]$ contains lane information in a traffic scenario, including lane position and lane type. For each line $l \in L$, it represents one lane line information at the current time step, containing one-hot encoding of the location of all points on the lane line and the type of the lane line.

\section{Approach }
In order to generate agents actions closer to the real world, we designed a model that integrates Diffusion and Transformer, which is referred to as TSDiT (Traffic Scene Diffusion Models With Transformers). TSDiT consists of three parts, firstly the DiT(Diffusion with Transformer) block which is used to generate agents latent, then an encoder which is used to extract the agents features and HD map features in the traffic scene, and finally a trajectory decoder which is used to output the future trajectories of the agents. The next sections present details of the these parts separately. The overview of the diffusion model is provided in Figure~\ref{fig:diff}.

\subsection{Diffusion Models}
\begin{figure}[h]
    \centering
    \includegraphics[width=0.8\textwidth]{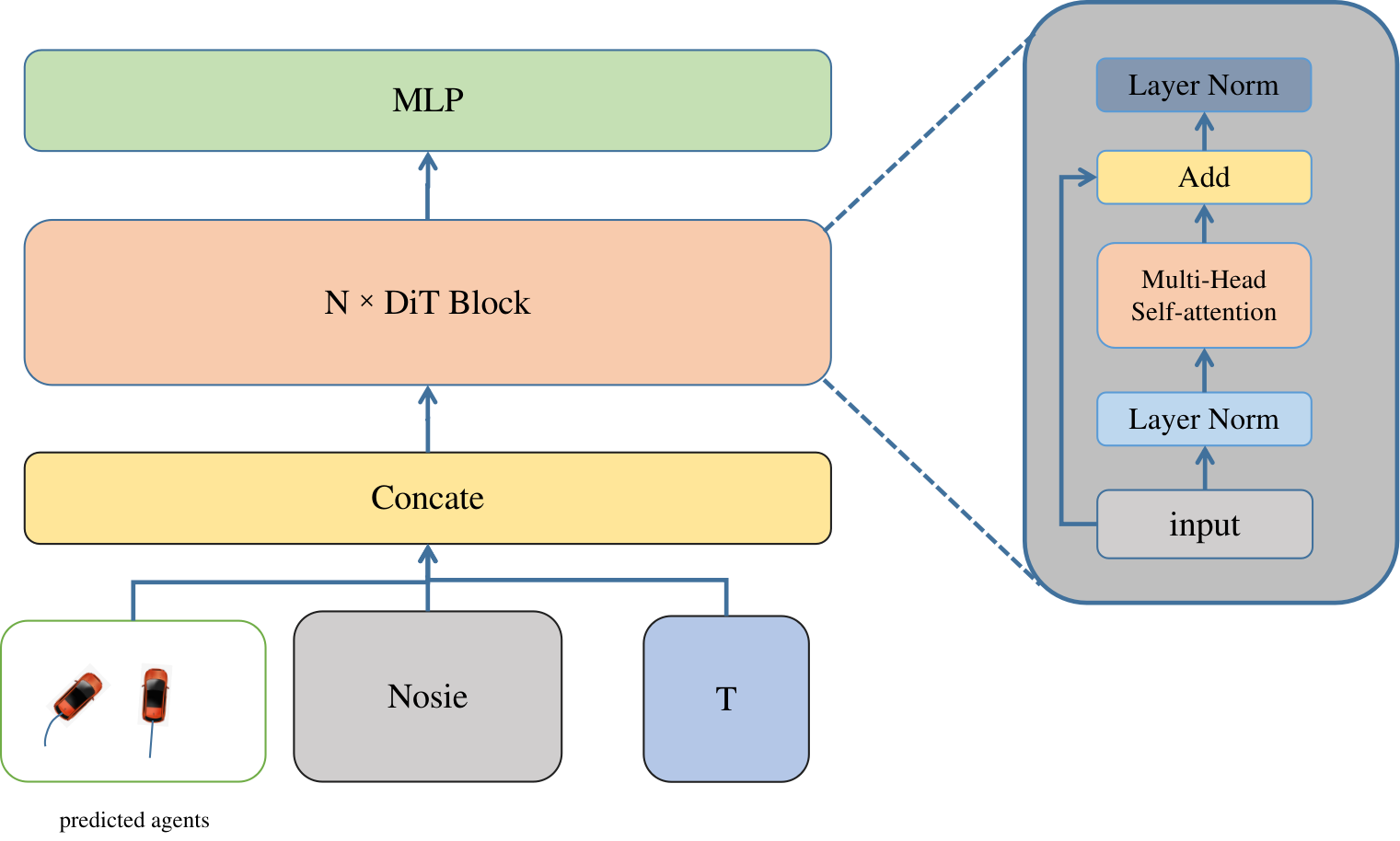}
    \caption{Overview of Diffusion with Transformers}
    \label{fig:diff}
\end{figure}

To increase the diversity of generated trajectories, we have integrated Denoising Diffusion Probabilistic Models (DDPM) in TSDiT. The main purpose of the diffusion model is to add "action latent" to the features of predicted agents. The main purpose of the diffusion model is to increase the uncertainty of agents' actions by adding "action latent" to the predicted agents' features. In addition, to improve the performance of the diffusion model, we use DiT blocks instead of Unet in DDPM, as shown in Figure~\ref{fig:diff}.

\begin{figure}[htbp]
    \centering
    \includegraphics[width=1.2\textwidth]{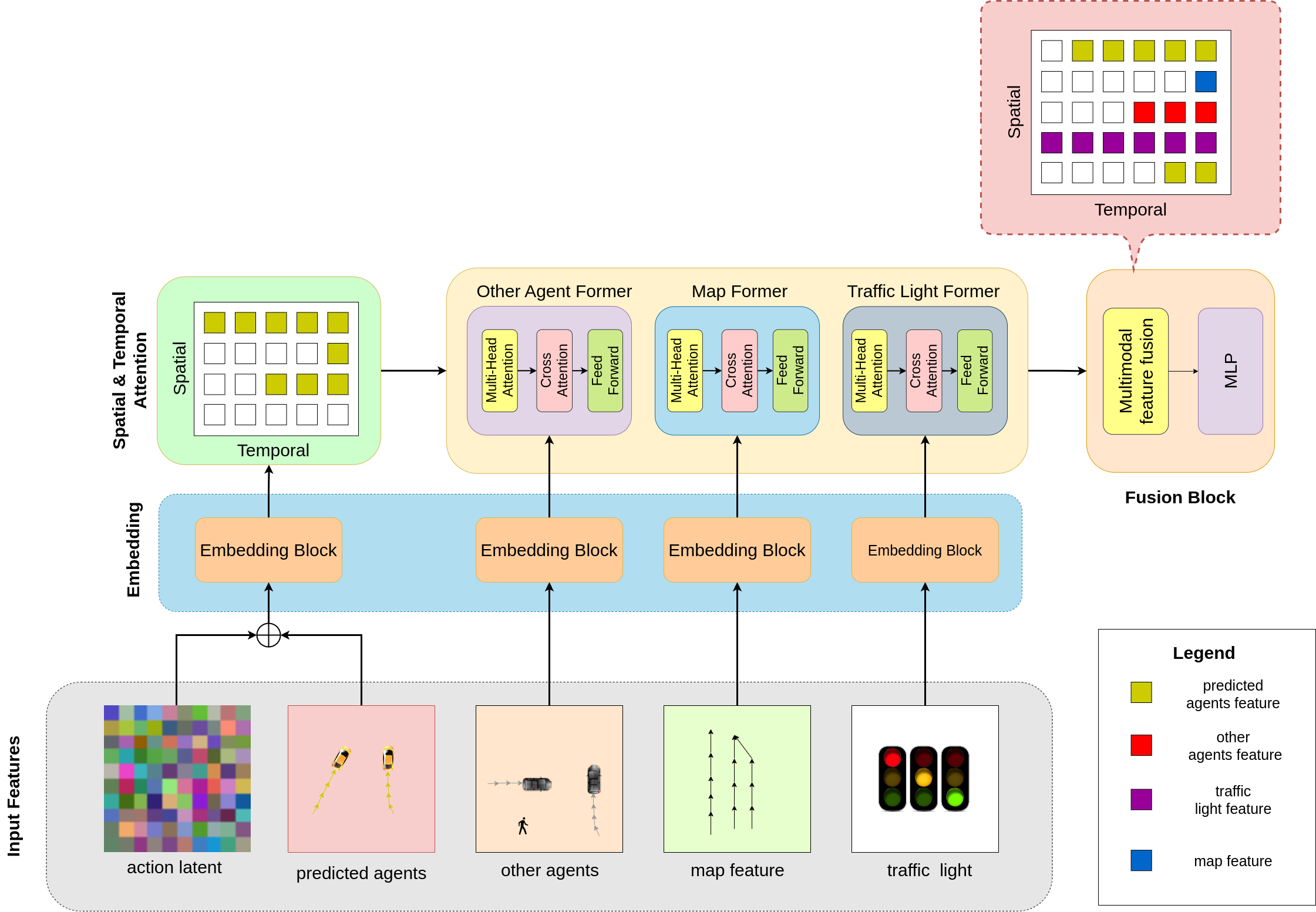}
    \caption{
        Overview of World-centric Encoders, which consists of four parts: (a) Embedding Blocks are used to encode the position and features of multimodal scene information (b) Spatial and Temporal Attention serves to encode the temporal and spatial information of predicted agents (c) Scene Formers model information from other modalities in the traffic scene in temporal and spatial dimensions (d) Fusion Blocks fuse the previous features and output the results to the trajectory encoder.
    }
    \label{fig:encoder}
\end{figure}

\subsection{Embedding Block}
The main purpose of Embedding Block is to encode the features (length, width, type) and location information of the agents in the traffic scenario, it can encode angets features globally and avoiding the transformation of the coordinate system to a Frenet coordinate system centred on individual angets, as in the case of trajectory prediction models. Embedding Block consists of two Multi-Layer Perceptrons(MLPs) Pose-Embedding(encode all positional features $\text{P}_i$ into a one-dimensional matrix) and Feature-Embedding(encode all agent features [$\text{height}, \text{width}, \text{type\textunderscore onehot}$] into a one-dimensional matrix). For each agent $i$ : 

\begin{align*}
E_p & = \phi_{p\text{-embedding}}[x_i, y_i]\\
\end{align*}
where ${p\text{-embedding}}$ is a linear layer, $E_p$ is the result of embedding and {$x_i$, $y_i$ are the position matrices of the agent.}

\begin{align*}
E_f & = \phi_{f\text{-embedding}}[A_w, A_h, A_{\text{type}}] \\
\end{align*}
where $\phi_{f\text{-embedding}}$ is a linear layer, {$A_w$ is the width of the agent.}, {$A_h$ is the height of the agent.}, $A_{\text{type}}$ is the type of the agent (needs to be processed into one-hot format) and $E_f$ is the result of embedding.

\begin{align*}
E_{\text{agent}} = \text{ReLU}(\text{LayrNorm}(\text{Concat}(E_p, E_f)))
\end{align*}

\subsection{Other Agent Former}
In order to help the model learn the association between all agents in the traffic scenario, we propose the Other Agent Former which consists of Multi-head Attention Block and Cross Attention Block. Firstly, we extract the features of predicted agents using multi-head attention block and then we extract the features of predicted agents using cross attention block. First, we use the multi-head attention block to extract the features of predicted agents and then, we use the cross-attention block to use the predicted agents features as the query matrix and the other agents features as the key and value matrix.

\textbf{Other Agent Former Self Attention:}

\begin{align*}
q_{p\text{-agent}} &= W^{Q \times \text{head}} E_{p\text{-agent}}, \quad k_{p\text{-agent}} = W^{K \times \text{head}} E_{p\text{-agent}}, \quad v_{p\text{-agent}} = W^{V \times \text{head}} E_{p\text{-agent}} \\
\label{eq1}
\end{align*}
where $W^Q$, $W^K$, $W^V$ are learnable parameters and $E_{p\text{-agent}}$ is the result of the predicted agent after the embedding block.
\begin{align*}
\propto_{p\text{-agent}} &= \text{Softmax}\left(\frac{q_{p\text{-agent}}^T}{\sqrt{d_k}} k_{p\text{-agent}}\right) \\
\text{Self}_{p\text{-agent}} &= \propto_{p\text{-agent}} v_{p\text{-agent}}
\end{align*}

After the features of predicted agents pass through the multi-head self-attention module and the cross-attention module with the features of other agents, we introduce the residual module to avoid model decay and the result of this module will inputted into HD Map Former.

\textbf{Cross Attention:}
\begin{align*}
q_{p\text{-agent}} &= W^Q E_{p\text{-agent}}, \quad k_{o\text{-agent}} = W^K E_{o\text{-agent}}, \quad v_{o\text{-agent}} = W^V E_{o\text{-agent}} \\
\propto_{p\text{-agent}} &= \text{Softmax}\left(\frac{q_{p\text{-agent}}^T}{\sqrt{d_k}} k_{o\text{-agent}}\right) \\
\text{Cross}_{p\text{-agent}} &= \propto_{p\text{-agent}} v_{o\text{-agent}}
\end{align*}
where $W^Q$, $W^K$, $W^V$ are learnable parameters, $E_{p\text{-agent}}$ is the result of the predicted agent after the embedding block and $E_{o\text{-agent}}$ is the result of the other agent after the embedding block.


\subsection{HD Map Former}
In order for the model to acquire the map information in the traffic scene, we introduce the HD Map Former, which consists of a Multi-head Attention Block and a Cross Attention Block. In HD Map Former, we use Bird's Eye View (BEV) maps and split them into different paths as map features. Firstly, we input the results of Other Agent Former into the Multi-head Self-Attention Block, and then we use the results of the Multi-head Self-Attention Block as the query matrix and the map features as the key and value matrix.

\textbf{Cross Attention:}
\begin{align*}
q_{p\text{-agent}} &= W^Q E_{p\text{-agent}}, \quad k_{\text{map}} = W^K E_{\text{map}}, \quad v_{\text{map}} = W^V E_{\text{map}} \\
\propto_{p\text{-agent}} &= \text{Softmax}\left(\frac{q_{p\text{-agent}}^T}{\sqrt{d_k}} k_{\text{map}}\right) \\
\text{Cross}_{p\text{-agent}} &= \propto_{p\text{-agent}} v_{\text{map}}
\end{align*}
where $W^Q$, $W^K$, $W^V$ are learnable parameters, $E_{p\text{-agent}}$ is the result of the predicted agent after the embedding block and $E_{\text{map}}$ is the result of HDMap after the embedding block. We consider the output for Cross-Attention as :
\begin{align*}
\text{Output}_{\text{cross-attention}} = \phi_{\text{output}}[\text{Cross}_{p\text{-agent}}] + E_{p\text{-agent}}
\end{align*}
After the features of predicted agents have passed through the multi-head self-attention module and the cross-attention module of HD Map features, the model has completed the feature extraction of the whole traffic scene, and then the results of HD Map Former will be inputted into the decoder.

\subsection{Trajectory Decoder}
After the agent features and HD map features in the traffic scene have been encoded, we introduce the Self-Attention Block aiming to summarise the above information and extract the important information. We then use a multi-layer perceptrons (MLPs) to decode the trajectories of the predected agents. To reduce the negative impact of positional differences between agents, we use the difference in the x-direction and the difference in the y-direction at each time step as model outputs and use them to calculate the agents' future trajectory, heading and speed. We compute the Trajectory as:
\begin{align*}
Traj_{agent} = Pos_{agent} + \sum_{i=0}^{t} [\Delta x, \Delta y]
\end{align*}

This formula represents the trajectory points of the agent (\( \text{{Traj}}_{\text{{agent}}} \)) over time (\(i = 0\) to \(t\)). Each point is determined by the initial position of the agent (\( \text{{Pos}}_{\text{{agent}}} \)) plus the incremental displacement \([\Delta x, \Delta y]\) for each time step (\(i\)).

\begin{align*}
\theta_{\text{agent}} = \arctan2(\Delta y, \Delta x)
\end{align*}

The heading (\( \theta_{\text{{agent}}} \)) is calculated using the arctangent function (\( \arctan2 \)) based on the differences in \(x\) and \(y\) coordinates (\( \Delta x, \Delta y \)).

\begin{align*}
\text{Speed}_{\text{agent}} = [\Delta x, \Delta y] \times \Delta t
\end{align*}

The speed (\( \text{{Speed}}_{\text{{agent}}} \)) is determined by the vector product of the displacement \([\Delta x, \Delta y]\) and the time increment (\( \Delta t \)).

\section{Training}

The total loss (\( \text{L}_{\text{total}} \)) during training is composed of various components:
\begin{align*}
L_{\text{total}} = L_{\text{Diffusion}} + L_{\text{W-ADE}} + L_{\text{FDE}} + \text{Huber}_{\text{Virtual-Trajectory}}
\end{align*}

The Weighted Average Displacement Error (\( \text{L}_{\text{W-ADE}} \)) is defined as:
\begin{align*}
L_{\text{W-ADE}} = \text{Weight}_{\text{time-step}} \times \text{ADE}
\end{align*}

The Final Displacement Error (\( \text{L}_{\text{FDE}} \)) is calculated as the distance between the predicted endpoint and the ground truth endpoint:
\begin{align*}
L_{\text{FDE}} = \text{Distance}(\text{endpoint}_{\text{pred}}, \text{endpoint}_{\text{ground-truth}})
\end{align*}

\begin{align*}
\text{Huber Loss}(y, f(x)) = 
\begin{cases}
  \frac{1}{2}(y - f(x))^2 & \text{for } |y - f(x)| \leq \delta \\
  \delta(|y - f(x)| - \frac{1}{2}\delta) & \text{otherwise}
\end{cases}
\end{align*}

Firstly, Mean Squared Error (MSE) Loss is utilized as the loss function for DiT, ensuring proximity of noise points estimated by DiT to the agent's action distribution. Additionally, Weighted Average Displacement Error (W-ADE) Loss is employed to fit trajectories with shorter time steps. For generating trajectory endpoints near the ground truth, Final Displacement Error (FDE) Loss is introduced to evaluate the deviation of the generated trajectory endpoints. The Virtual Trajectory Loss is a Huber loss between the predicted future trajectory and a virtual trajectory, aiming to reduce trajectory curvature for smoother turning curves. During training, the curvature of trajectories significantly increases when agents turn at intersections. To address this, we preprocess the ground truth, using cubic interpolation to generate a "virtual trajectory", and employ Huber Loss to assess the error between the "virtual trajectory" and the generated trajectory. This loss function is often used in regression problems and is less sensitive to outliers than traditional MSE, it provides a balance between the robustness of MAE and the smoothness of MSE.

\section{Experiment and Result Analysis }

\subsection{Dataset}
We train and evaluate the traffic scene generation model using the Waymo Motion Prediction dataset~\cite{montali2023waymo}, which provides agents' motion trajectories and high-definition map data. The dataset contains more than 400000 real driving scenarios divided into training, validation and test sets. All traffic scenarios are 9 seconds long and they are 9-second sequences sampled at 10 Hz. In the test scenario, only the first 1 second of the trajectory is publicly available, whereas the simulated driver challenge requires the generation of agents' motion trajectories for the following 8 seconds.
\subsection{Metrics}
To evaluate whether the data generated by our model matches the original data, We use the mean
maximum discrepancy (MMD) to evaluate the similarity between the test data and the generated data, including: positional similarity, velocity similarity, angular similarity of orientation. In addition, we introduced ade and fde to evaluate the distributional differences between the generated trajectories and the ground truth. However, the angle and speed of MVTA-generated agents are more similar to ground truth. Although the speed MMD and heading MMD of TSDiT are slightly higher than that of MVTA, the ADE and FDE of TSDiT are significantly lower than that of MVTA.Therefore, the performance of TSDiT is more balanced.

\subsection{Comparison Results}
\begin{table}[htbp]
    \centering
    \caption{Performance Metrics of Various Methods}
    \begin{tabular}{cccccc}
        \toprule
        & \textbf{Method} & \textbf{ADE} & \textbf{FDE} & \textbf{Speed MMD} & \textbf{Heading MMD} \\
        \midrule
    & MTR++ ~\cite{shi2023mtr++}& 0.691 & 1.75 & 0.473 & 0.311 \\
        & WayFormer~\cite{nayakanti2023wayformer} & 0.985 & 2.31 & 0.582 & 0.276 \\
        & MVTA~\cite{wang2023multiverse} & 0.822 & 1.86 & 0.439 & 0.248 \\
        & MULTIPATH++~\cite{varadarajan2022multipath++} & 0.96 & 1.763 & 0.438 & 0.273 \\
        & TSDiT & 0.684 & 1.792 & 0.452 & 0.261 \\
        \bottomrule
    \end{tabular}
    \label{tab:my_label}
\end{table}

We compared our model to state-of-the-art models on the Sim Agents test set. Similar to the results of Sim Agents~\cite{montali2023waymo}, the ADE and FDE of MTR++~\cite{shi2023mtr++} were significantly lower than the other methods in our experiments, and the ADE of TSDiT was slightly better than that of MTR++, but the FDE was increased compared to MTR++.

We present qualitative results for TSDiT on the Sim Agents validation set and show the locations of the other agents in the traffic scenario separately, the trajectories and locations of the predicted agents. Overall, the experiment results show that TSDiT can generate the agents' turning at junctions, but the trajectory is not smooth enough. Through the ablation experiments we found that due to our introduction of the Diffusion Model, which led to an increase in the diversity of actions generated by the model also increased the randomness of the trajectory resulting in an unsmooth trajectory. 


\subsection{Ablation Studies}
\subsubsection{Importance of Each Block} 
We tested the importance of individual modules to the model by adding or removing a number of blocks, each of which improves prediction performance to some extent. First, without Other Agent Former, the model is unable to capture the historical motion of nearby agents, which results in the collision of generated trajectories with nearby agents. Then, the HD Map Former is also an important module in the model, and our attempts to remove it in our experiments resulted in generated trajectories that did not meet the lane constraints. We also note that increasing the number of layers and dimensions of this module can further improve the performance, but for the sake of efficiency, we do not increase the number of layers of each module. In addition, Embedding-Block can significantly improve the model's understanding of position and agent attributes.
\subsubsection{Ablation Studies on Decoder}
We evaluated the contribution of the multi-head attention block in Decoder to the model, and we introduced it with the main purpose of extracting important information from the Encoder results. Removing the multi-head attention block from the decoder would result in a slower model convergence because of the reduced efficiency in extracting the Encoder information. Similarly, adding multi-head attention blocks can improve the model performance, but it will significantly increase the consumption of computer resources.

\begin{table}[htbp]
  \centering
  \captionsetup{skip=5pt} 
  \caption{Abalative study of TSDiT}
  \scalebox{1}{
    \setlength\tabcolsep{2pt}
    \begin{tabular}{ccccc|cc}
      \toprule
      \textbf{Base} & \textbf{\makecell[c]{Other Agent \\Former}} & \textbf{\makecell[c]{HD Map \\Former}} & \textbf{\makecell[c]{Self Attention \\in Decoder}} &  &
      \textbf{FDE}\ensuremath{\downarrow} & \textbf{ADE}\ensuremath{\downarrow}  \\
      \midrule
      \checkmark &            & \checkmark & \checkmark &  & 2.090 & 0.809 \\
      \checkmark & \checkmark &            & \checkmark &  & 3.040 & 1.688 \\
      \checkmark & \checkmark & \checkmark &            &  & 2.021 & 0.984 \\
      \checkmark & \checkmark & \checkmark & \checkmark &  &  \textbf{1.792} & \textbf{0.684} \\
      \bottomrule
    \end{tabular}
    }
\label{table:ablation}
\end{table}


\section{Conclusion }

In this paper, we proposed a novel approach to trajectory generation for autonomous driving by leveraging the synergies between Diffusion models and Transformers. 

We find that the fusion of action latent, historical trajectories, and bird's eye view (BEV) information within distinct transformer blocks constitutes a key innovation in our methodology. This integration aims to capture complex interactions within the traffic scene. Subsequently, a trajectory decoder is employed to predict future trajectories for agents in the autonomous driving environment.

Our method has better performance in generating smooth turning trajectories, showcasing an improved ability to model complex steering patterns. Experimental results confirm the effectiveness of our approach, demonstrating its capacity to produce realistic and diverse trajectories.

\small{
\bibliographystyle{IEEEtran}
\bibliography{mylib.bib} }

\begin{thebibliography}{10}
\providecommand{\url}[1]{#1}
\csname url@samestyle\endcsname
\providecommand{\newblock}{\relax}
\providecommand{\bibinfo}[2]{#2}
\providecommand{\BIBentrySTDinterwordspacing}{\spaceskip=0pt\relax}
\providecommand{\BIBentryALTinterwordstretchfactor}{4}
\providecommand{\BIBentryALTinterwordspacing}{\spaceskip=\fontdimen2\font plus
\BIBentryALTinterwordstretchfactor\fontdimen3\font minus \fontdimen4\font\relax}
\providecommand{\BIBforeignlanguage}[2]{{%
\expandafter\ifx\csname l@#1\endcsname\relax
\typeout{** WARNING: IEEEtran.bst: No hyphenation pattern has been}%
\typeout{** loaded for the language `#1'. Using the pattern for}%
\typeout{** the default language instead.}%
\else
\language=\csname l@#1\endcsname
\fi
#2}}
\providecommand{\BIBdecl}{\relax}
\BIBdecl

\bibitem{dosovitskiy2017carla}
A.~Dosovitskiy, G.~Ros, F.~Codevilla, A.~Lopez, and V.~Koltun, ``Carla: An open urban driving simulator,'' in \emph{Conference on robot learning}.\hskip 1em plus 0.5em minus 0.4em\relax PMLR, 2017, pp. 1--16.

\bibitem{behrisch2011sumo}
M.~Behrisch, L.~Bieker, J.~Erdmann, and D.~Krajzewicz, ``Sumo--simulation of urban mobility: an overview,'' in \emph{Proceedings of SIMUL 2011, The Third International Conference on Advances in System Simulation}.\hskip 1em plus 0.5em minus 0.4em\relax ThinkMind, 2011.

\bibitem{yang2022diffusion}
L.~Yang, Z.~Zhang, Y.~Song, S.~Hong, R.~Xu, Y.~Zhao, W.~Zhang, B.~Cui, and M.-H. Yang, ``Diffusion models: A comprehensive survey of methods and applications,'' \emph{ACM Computing Surveys}, 2022.

\bibitem{montali2023waymo}
N.~Montali, J.~Lambert, P.~Mougin, A.~Kuefler, N.~Rhinehart, M.~Li, C.~Gulino, T.~Emrich, Z.~Yang, S.~Whiteson \emph{et~al.}, ``The waymo open sim agents challenge,'' \emph{arXiv preprint arXiv:2305.12032}, 2023.

\bibitem{pronovost2023generating}
E.~Pronovost, K.~Wang, and N.~Roy, ``Generating driving scenes with diffusion,'' \emph{arXiv preprint arXiv:2305.18452}, 2023.

\bibitem{zhong2023guided}
Z.~Zhong, D.~Rempe, D.~Xu, Y.~Chen, S.~Veer, T.~Che, B.~Ray, and M.~Pavone, ``Guided conditional diffusion for controllable traffic simulation,'' in \emph{2023 IEEE International Conference on Robotics and Automation (ICRA)}.\hskip 1em plus 0.5em minus 0.4em\relax IEEE, 2023, pp. 3560--3566.

\bibitem{zhong2023language}
Z.~Zhong, D.~Rempe, Y.~Chen, B.~Ivanovic, Y.~Cao, D.~Xu, M.~Pavone, and B.~Ray, ``Language-guided traffic simulation via scene-level diffusion,'' \emph{arXiv preprint arXiv:2306.06344}, 2023.

\bibitem{nayakanti2023wayformer}
N.~Nayakanti, R.~Al-Rfou, A.~Zhou, K.~Goel, K.~S. Refaat, and B.~Sapp, ``Wayformer: Motion forecasting via simple \& efficient attention networks,'' in \emph{2023 IEEE International Conference on Robotics and Automation (ICRA)}.\hskip 1em plus 0.5em minus 0.4em\relax IEEE, 2023, pp. 2980--2987.

\bibitem{shi2023mtr++}
S.~Shi, L.~Jiang, D.~Dai, and B.~Schiele, ``Mtr++: Multi-agent motion prediction with symmetric scene modeling and guided intention querying,'' 2023.

\bibitem{wang2023multiverse}
Y.~Wang, T.~Zhao, and F.~Yi, ``Multiverse transformer: 1st place solution for waymo open sim agents challenge 2023,'' \emph{arXiv preprint arXiv:2306.11868}, 2023.

\bibitem{varadarajan2022multipath++}
B.~Varadarajan, A.~Hefny, A.~Srivastava, K.~S. Refaat, N.~Nayakanti, A.~Cornman, K.~Chen, B.~Douillard, C.~P. Lam, D.~Anguelov \emph{et~al.}, ``Multipath++: Efficient information fusion and trajectory aggregation for behavior prediction,'' in \emph{2022 International Conference on Robotics and Automation (ICRA)}.\hskip 1em plus 0.5em minus 0.4em\relax IEEE, 2022, pp. 7814--7821.

\end{thebibliography}

\end{document}